\documentclass[runningheads]{llncs}

\usepackage[T1]{fontenc}
\usepackage{graphicx}
\usepackage{amsmath,amssymb}
\usepackage{booktabs}
\usepackage{xcolor}
\usepackage{subcaption}
\usepackage{microtype}
\usepackage{hyperref}

\newcommand{\dt}{\Delta t}
\newcommand{\dttrain}{\dt_{\mathrm{train}}}
\newcommand{\dteval}{\dt_{\mathrm{eval}}}

\newcommand{\Tperiod}{T_{\mathrm{period}}}

\begin{document}

\title{Unlocking Temporal Generalization in Hamiltonian Video Dynamics Models}

\author{Eli Laird\inst{1}\orcidID{0000-0002-0668-8745} \and
Corey Clark\inst{1}\orcidID{0000-0003-0129-9270}}

\authorrunning{E.~Laird and C.~Clark}

\institute{Department of Computer Science, Southern Methodist University,\\
Dallas, TX, USA\\
\email{\{ejlaird,coreyc\}@smu.edu}}

\maketitle

\begin{abstract}

World models are typically trained to predict discrete-time physical dynamics with a fixed step size baked into the model weights, preventing prediction at variable temporal resolutions. This matters for hierarchical planning, sim-to-real transfer, and scientific or game-engine applications that must query the same dynamics at multiple timescales. Hamiltonian Generative Networks (HGN) offer a principled path forward, grounding predictions in a continuous-time energy function that is, in principle, independent of the observation frame rate. In practice, however, their temporal generalization breaks down in non-conservative settings. We show that in externally forced, dissipative environments, HGN rollouts at step sizes beyond the training regime fail due to distinct failure modes, including latent magnitude growth driven by an unconstrained action-force map, and global truncation error accumulation from an under-resolved integrator. We identify a targeted fix for each mechanism and demonstrate stable dynamics prediction at temporal resolutions well outside the training distribution. In a detailed analysis, we recommend several strategies for enabling temporal generalization in continuous-time video generation.

\keywords{Physics-aware Video Generation \and Physics-informed Neural Networks \and
Temporal Generalization \and Hamiltonian Dynamics}
\end{abstract}

\section{Introduction}
\label{sec:intro}

Most world model architectures predict physical dynamics on a discrete time grid with a fixed step size at training time. The dominant paradigm in world modeling is a discrete-time transition $z_{t+1} = f(z_t, a_t)$, the structure underlying the early World Model architectures~\cite{watter2015embed,ha2018world,ha2018reccurentworld}, the Dreamer series~\cite{hafner2020dream,hafner2021mastering,hafner2025mastering}, the JEPA-based models \cite{lecun2022path,bar2024navigation,zhou2025dino,maes2026leworldmodel}, and others~\cite{alonso2024diffusion,Valevski2024DiffusionMA,gumbsch2024thick,po2025long}. These models learn capable predictors at their training rate but have no mechanism to generalize across temporal resolutions, which matters for hierarchical planning (coarse and fine time scales)~\cite{sutton1999between,hafner2022director,Zhang2026HierarchicalPW}, sim-to-real transfer (simulation and control rates rarely match)~\cite{peng2018simtoreal,tallec2019making}, and scientific simulation engines that must query the same dynamics at multiple timescales~\cite{brandstetter2022message}.

Neural ODEs~\cite{chen2018neural,rubanova2019latent} address the fixed-step constraint by parameterizing dynamics as a vector field $\frac{dz}{dt} = f(z)$ and integrating at any desired step size. But the learned field is an unconstrained network, which without physical inductive biases may not respect conservation laws or symplectic structure, destabilizing long-horizon rollouts.

Hamiltonian neural networks provide exactly this conservative and symplectic structure \cite{greydanus2019hamiltonian}. The Hamiltonian is a scalar energy function $H(q,p)$ with gradients that define the equations of motion and guarantee energy conservation, time reversibility, and symplectic flow. Hamiltonian Generative Networks (HGN)~\cite{toth2020hamiltonian} demonstrated plausible visual rollouts at double and half the training rate from a single model, as well as time-reversed dynamics. Yet in non-conservative settings this temporal generalization degrades, and its limits have not been systematically characterized.

We study temporal generalization in Hamiltonian Generative Networks through the lens of interpolation and extrapolation, predicting at step sizes, respectively, smaller and larger than training. Interpolation generalizes naturally, since the symplectic integrator refines any target step without retraining; extrapolation, where the step exceeds anything seen during training, is more fragile and is the focus of this work (Section~\ref{sec:theory}).

\begin{figure}[t]
\includegraphics[width=1.0\columnwidth]{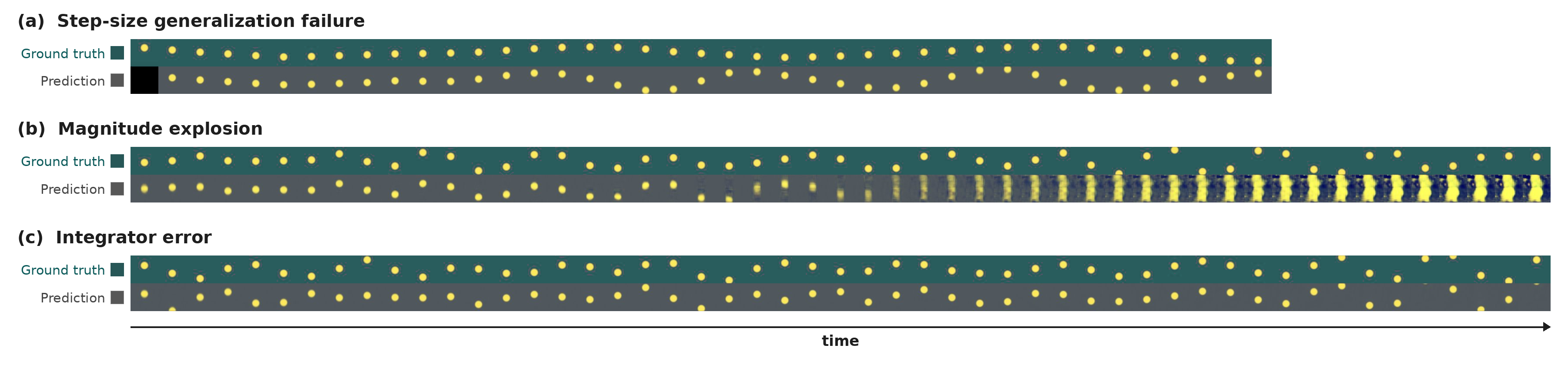}
\caption{Three views of how dynamics rollouts can fail at evaluation step sizes beyond the training step. In each panel the top strip (teal) is the ground-truth frame sequence and the bottom strip (gray) is the model's open-loop prediction. \textbf{(a)}~\emph{Step-size generalization failure}: past the training step the predicted rollout defaults to the memorized dynamics at the training step size. \textbf{(b)}~\emph{Magnitude explosion}: an unconstrained integrator injects energy causing the rendered mass smears into high-amplitude artifacts. \textbf{(c)}~\emph{Integrator error}: with the energy-explosion constrained, the rollout stays bounded but accumulates global truncation error from the under-resolved integrator, drifting out of phase with the ground truth.}
\label{fig:failure_modes}
\end{figure}

We extend HGN with port-Hamiltonian structure to handle externally forced, dissipative environments and identify multiple failure modes where rollouts fail beyond the training regime, including exponential latent magnitude growth driven by an unconstrained action-force map, and global truncation error accumulation from an under-resolved integrator (Fig.~\ref{fig:failure_modes}). We provide a targeted fix for each and suggest a strategy for learning visual continuous-time dynamics at multiple temporal resolutions.

\section{Related Work}
\label{sec:related}

\subsection{World Models}

Most world models predict environment dynamics as a discrete-time latent transition $z_{t+1} = f(z_t, a_t)$ learned at a single training frame rate. Early latent dynamics models established this template directly from pixels. Embed to Control~\cite{watter2015embed} learned a locally linear latent transition for control from raw images, and the World Models of Ha and Schmidhuber~\cite{ha2018world,ha2018reccurentworld} paired a variational autoencoder with a recurrent mixture-density network that rolls the latent state forward one step at a time. The Dreamer line scaled this recipe. Dreamer~\cite{hafner2020dream} learns behaviors by latent imagination inside a recurrent state-space model, DreamerV2~\cite{hafner2021mastering} replaced the continuous latent with discrete categorical codes, and DreamerV3~\cite{hafner2025mastering} stabilized the approach across diverse domains. In every case the recurrent update advances the state by exactly one fixed environment step, with no notion of the underlying continuous time.

A second family predicts in a learned representation space rather than in pixels. The joint-embedding predictive architecture (JEPA) proposed by LeCun~\cite{lecun2022path}, with image-level instantiations such as I-JEPA~\cite{assran2023self} and the collapse-free LeJEPA~\cite{balestriero2025lejepa}, motivates predicting future embeddings instead of reconstructing observations. Built on top of these JEPA encoders, DINO-WM~\cite{zhou2025dino} learns next-step dynamics over frozen DINO features to enable zero-shot planning, Navigation World Models~\cite{bar2024navigation} predict future egocentric observations conditioned on navigation actions, and LeWorldModel~\cite{maes2026leworldmodel} trains a JEPA next-embedding predictor end-to-end from pixels. These predictors are likewise autoregressive over a fixed step and cannot be queried at an arbitrary temporal resolution.

Generative video models form a third family: DIAMOND~\cite{alonso2024diffusion} learns a diffusion world model that preserves fine visual detail for agent training, GameNGen~\cite{Valevski2024DiffusionMA} turns a diffusion next-frame predictor into a real-time neural game engine, and Long-Context State-Space Video World Models~\cite{po2025long} adopt state-space backbones to extend the memory of autoregressive frame prediction. A few models reason across multiple time scales, but only over integer multiples of the base step. THICK~\cite{gumbsch2024thick} learns a hierarchy in which a high level predicts sparse context-change events above a discrete low-level latent dynamics. Across all of these families the predictor is bound to the temporal resolution of its training data, advancing by a fixed step rather than integrating a continuous-time vector field.

\subsection{Neural ODEs and Continuous-time Dynamics in State Space}

A parallel line of work, focused on Neural ODEs, has embedded physics into learned dynamics through architectural inductive biases. Hamiltonian Neural Networks~\cite{greydanus2019hamiltonian} learned a scalar Hamiltonian $H(q,p)$ from state observations, demonstrating energy conservation on simple systems. Lagrangian Neural Networks~\cite{cranmer2020lagrangian} took the dual approach in position-velocity space. Neural ODEs~\cite{chen2018neural} parameterized continuous-time dynamics via ODE solvers, making the timestep a free parameter, and Latent ODEs~\cite{rubanova2019latent} extended this to irregularly-sampled observations in latent space. Port-Hamiltonian neural networks~\cite{desai2021port} built on the port-Hamiltonian formalism~\cite{vanderschaft2004port} to handle dissipative systems with external forcing. All of these methods operate in known low-dimensional state spaces, while our model embeds the port-Hamiltonian structure within a pixel-to-pixel generative pipeline, introducing the additional challenge of jointly learning the visual state representation and the dynamics.
 
\subsection{Hamiltonian Generative Networks}
 
The approach most directly comparable to ours is Hamiltonian Generative Networks (HGN)~\cite{toth2020hamiltonian}, which first learned the Hamiltonian dynamics end-to-end from pixel observations. HGN infers a latent phase-space state, integrates a learned separable Hamiltonian with a symplectic leapfrog integrator, and decodes images from latent ``position'' coordinates. The model is trained with a temporally extended ELBO objective~\cite{kingma2013auto} and demonstrated forward, backward, $2\times$, and $\frac{1}{2}\times$ speed rollouts from a single model by changing the magnitude of the integration step. Our work extends HGN with a port-Hamiltonian external force (action) injection and characterizes when and why temporal generalization breaks down at larger step-size ratios.

\section{Interpolation and Extrapolation}
\label{sec:theory}

\begin{figure}[t!]
\centering
\includegraphics[width=0.9\textwidth]{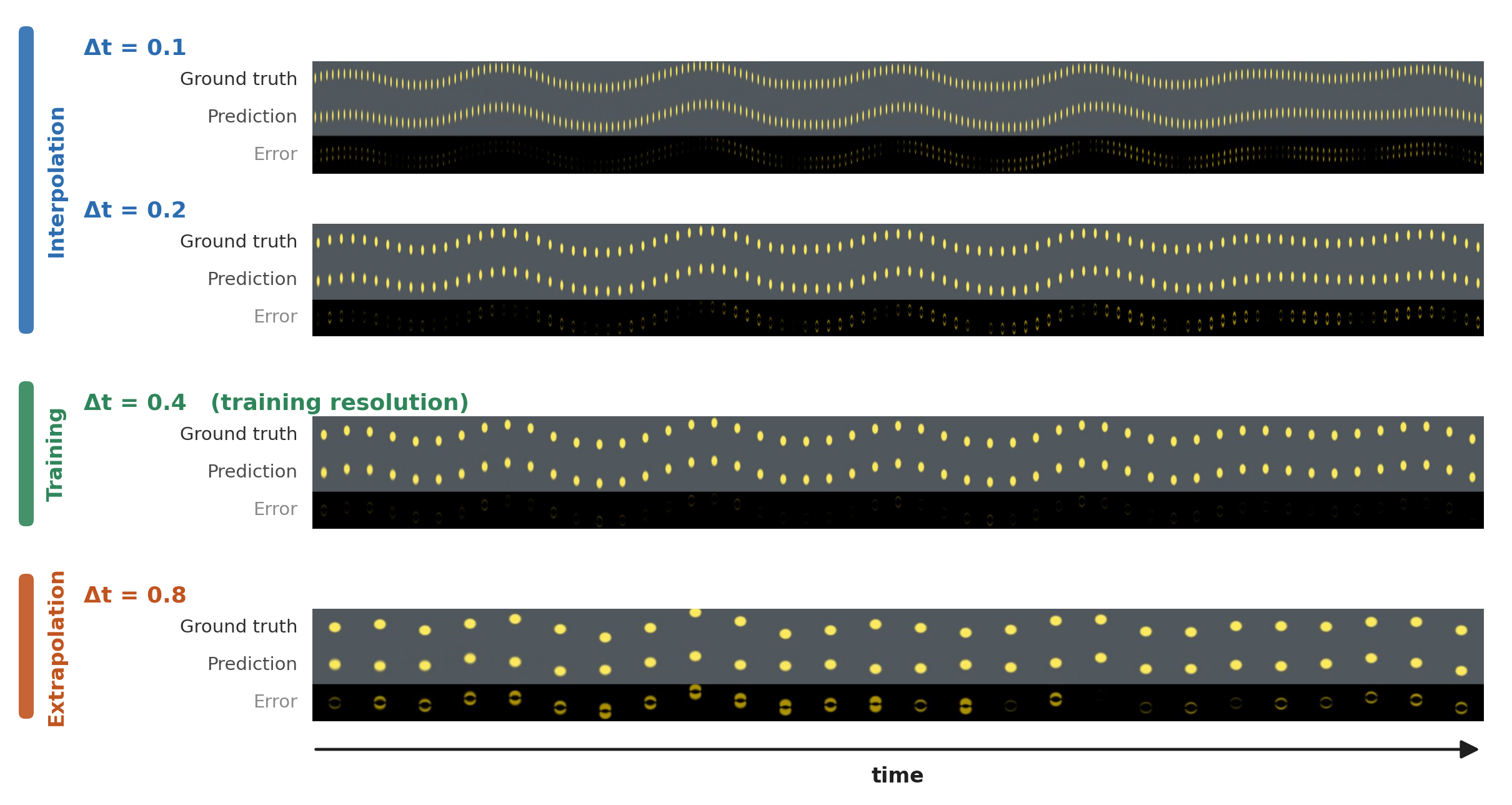}
\caption{Temporal generalization of the baseline port-HGN. Each block renders the same underlying oscillator trajectory sampled at a different evaluation step size $\dteval$, shown as ground truth (top), prediction (middle), and prediction error (bottom). Below the training step the model interpolates accurately ($\dteval \in \{0.1, 0.2\}$) and it reproduces the training resolution ($\dteval = \dttrain = 0.4$), with error rows that remain near-black. In extrapolation ($\dteval = 0.8$) the error row develops pronounced residuals as the prediction drifts out of phase.}
\label{fig:interp_extrap}
\end{figure}

Temporal generalization has two directions that test fundamentally different things about what a model has learned.
\textit{Interpolation} refers to evaluation at step sizes smaller than the training step. In this setting, the model samples its learned trajectory more densely, and the symplectic integrator can decompose any target step into finer substeps without retraining. Strong interpolation performance, however, does not distinguish a model that learned continuous-time physics from one that memorized a smooth curve.
\textit{Extrapolation} refers to evaluation at step sizes larger than training, where the integration step exceeds anything the model encountered during learning. A model that learned a genuine continuous-time vector field should fail for numerical reasons including the solver step being too large and producing truncation error that finer integration can remove. A model that memorized a fixed-step transition map should fail for representational reasons. These fixed-step models simply reproduce the dynamics at the training step size, and refining the integration cannot help. This asymmetry makes extrapolation a useful test for the continuous-time structure and is the focus of the analysis that follows.

\section{Methods}
\label{sec:model}

Our model follows the HGN design~\cite{toth2020hamiltonian} in which an encoder infers the dynamical state from pixels, a separable Hamiltonian network integrates the dynamics in an abstract phase space, and a decoder reconstructs images from the integrated state. We extend this with port-Hamiltonian structure, adding energy dissipation and action forcing.

\textbf{Encoder and Decoder.} A convolutional network with four residual blocks and stride-2 downsampling (64 channels) followed by a 2-layer MLP maps each 64$\times$64$\times$3 frame to a $d{=}64$ dimensional latent vector. The latent is partitioned equally into abstract position~$q$ and momentum~$p$. A convolutional upsampling network reconstructs images from position coordinates only: $\hat{x}_t = D_\rho(q_t)$, following Toth et al.~\cite{toth2020hamiltonian}.

\textbf{Port-Hamiltonian predictor.} Given state $(q, p)$ and action~$a$, the predictor integrates the port-Hamiltonian equations
\begin{equation}
\dot{q} = \frac{\partial H}{\partial p}, \qquad \dot{p} = -\frac{\partial H}{\partial q} - \gamma\,\frac{\partial H}{\partial p} + G(a),
\label{eq:port_ham}
\end{equation}
where $H(q,p) = T(p) + V(q)$ is a learned separable Hamiltonian with $T$ and $V$ each parameterized as scalar-output MLPs with Softplus activations, $\gamma \geq 0$ is a learned scalar damping coefficient, and $G(a)$ is a learned action-force map. 

\textbf{Leapfrog Integrator.} The physics integrator is a modified leapfrog~\cite{Verlet1967ComputerO} that folds the damping and forcing symmetrically into both momentum half-steps:
\begin{equation}
\begin{aligned}
p_{\mathrm{half}} &= p_t + \tfrac{\dt}{2}\!\left(-\frac{\partial H}{\partial q}\bigg|_t - \gamma\,\frac{\partial H}{\partial p}\bigg|_t + G(a)\right) \\
q_{t+1} &= q_t + \dt\,\frac{\partial H}{\partial p}\bigg|_{\mathrm{half}} \\
p_{t+1} &= p_{\mathrm{half}} + \tfrac{\dt}{2}\!\left(-\frac{\partial H}{\partial q}\bigg|_{t+1} - \gamma\,\frac{\partial H}{\partial p}\bigg|_{\mathrm{half}} + G(a)\right)
\end{aligned}
\label{eq:leapfrog}
\end{equation}
The action force $G(a)$ is held constant across both half-steps, so the per-step impulse is exactly $\dt \cdot G$ regardless of substepping level. The leapfrog integrator is conditionally stable, meaning, for an oscillator of frequency $\omega$, the update is stable only for $\dt < 2/\omega$.

\textbf{Objective.} All models are trained with the variational objective of Toth et al.~\cite{toth2020hamiltonian}. The encoder maps the context frames to a posterior $q_\phi(z \mid x_{0:T})$ over the initial latent state. A reparameterized sample is then projected to the initial phase-space state $(q_0, p_0)$, which is then rolled forward for the full horizon by the integrator without re-encoding or teacher forcing. Every predicted position $q_t$ in the trajectory is decoded independently, and the loss is
\begin{equation}
\mathcal{L}(\phi,\theta) = \frac{1}{T+1}\sum_{t=0}^{T}\mathbb{E}_{q_\phi(z \mid x_{0:T})}\!\left[\log p_{\theta}(x_t \mid q_t)\right] - \beta_{\mathrm{KL}} \cdot KL\!\left(q_\phi(z \mid x_{0:T}) \,\|\, p(z)\right),
\label{eq:elbo}
\end{equation}
where $\phi$ denotes the encoder parameters, $\theta$ the dynamics and decoder parameters, the first term is the per-frame reconstruction likelihood averaged over the rollout horizon, and the second is the KL divergence of the posterior against an isotropic unit Gaussian prior with $\beta_{\mathrm{KL}} = 1$. Because the entire trajectory is generated from a single initial encode, the reconstruction term is an open-loop rollout loss, i.e., the error at frame $t$ reflects $t$ accumulated integration steps. The decoder operates on position coordinates only which means momentum drives the dynamics but is never directly decoded. All models are optimized with Adam (learning rate $5 \times 10^{-4}$ to $5 \times 10^{-5}$, cosine schedule, and trained to convergence).

\textbf{Inference-time substepping.} To discriminate integrator error from representational failure, we introduce a inference-time substepping protocol. Given an observation step~$\dt$, we split each step into $N$ integration substeps of size $\dt/N$, applying the action force at each substep. The total action impulse is preserved ($N \cdot (\dt/N) \cdot G = \dt \cdot G$), so substepping affects only the integration resolution and not the action magnitude. The encoder, learned Hamiltonian, and inferred initial state are identical across all substepping conditions, therefore, any performance change is attributable to the reduction in truncation error.

\section{Experimental Setup}
\label{sec:setup}
 
\textbf{Environment.} We evaluate on a forced damped mass-spring oscillator rendered as 64$\times$64 RGB pixel observations, with mass $m = 0.5$, spring constant $k = 2.0$, and damping coefficient $c = 0.1$, giving natural frequency $\omega = 2$\,rad/s and period $\Tperiod \approx 3.14$ time units. The system is driven by three uniformly sampled discrete actions (up force, no force, down force). Initial states are sampled on energy contours with radius in $[0.5, 1.8]$ following \cite{toth2020hamiltonian}. The dataset contains 50{,}000 sequences of 50 frames at $\dttrain = 0.4$ (2.5\,Hz sampling).
 
\textbf{Evaluation.} We evaluate the oscillator at $\dteval \in \{0.05, 0.1, 0.15, 0.2, 0.3, \\ 0.4, 0.5, 0.6, 0.8, 1.0, 1.5\}$. for each evaluation trajectory we sample a single initial state and a reference action sequence at the training step size, then render the same dynamical trajectory at every $\dteval$ by integrating from the shared initial state at that step size, with the reference actions resampled so the per-action wall-clock impulses are preserved. The temporal horizon is the same across $\dteval$ values; what varies is the number of integration observation steps that cover it. For the substepping experiments, each observation step is split into $N \in \{1, 2, 4\}$ integration substeps with the action impulse preserved.
 
\textbf{Metrics.} We evaluate prediction quality in pixel space over the full open-loop rollout, reporting mean absolute error (MAE), PSNR \cite{hore2010image}, SSIM \cite{wang2004image}, and the perceptual LPIPS distance~\cite{zhang2018unreasonable} between predicted and ground-truth frames at every evaluation step size. These visual metrics primarily measure the differences in visual attributes between two images and, therefore, are not a perfect measure for trajectory error of physical dynamics. Since the physical state is unknown for the predicted trajectories in this setting, we rely on visual reconstructions and aggregated metrics across the full rollout horizon.

\section{Failure Decomposition}
\label{sec:decomposition}

When we evaluate the trained port-Hamiltonian model in the extrapolation regime, two visually distinct failure modes appear, each with a different cause and a different fix (Fig.~\ref{fig:failure_modes}).
 
\subsection{Energy Magnitude Growth}
\label{sec:result1}
 
With an unconstrained action map $G(a)$, predicted latents diverge in amplitude for step sizes beyond training, producing high-frequency artifacts in the decoded frames (see Fig.~\ref{fig:failure_modes}).
 
At large step sizes the numerical solver overshoots, advancing the state further per step than the true dynamics warrant. An unconstrained force map exploits this overshoot, injecting energy along the directions where the solver is least stable. Over multiple steps the injection compounds, producing the observed amplitude blow-up. From a spectral perspective, the action impulse has eigenvalues exceeding unity along certain phase-space directions at large $\dt$, and an unbounded $G$ compounds directly into those directions.
 
Applying spectral normalization~\cite{miyato2018spectral} to $G$ bounds its norm to 1, capping per-step energy injection. Action energy then enters linearly over the rollout and the amplitude blow-up is eliminated across the full tested $\dt$ range. We refer to this spectrally-normed variant as the \emph{bounded} port-HGN and use it in all experiments that follow.
 
\subsection{Phase Drift Accumulation}
\label{sec:result2}
 
\begin{figure}[t]
\centering
\includegraphics[width=\textwidth]{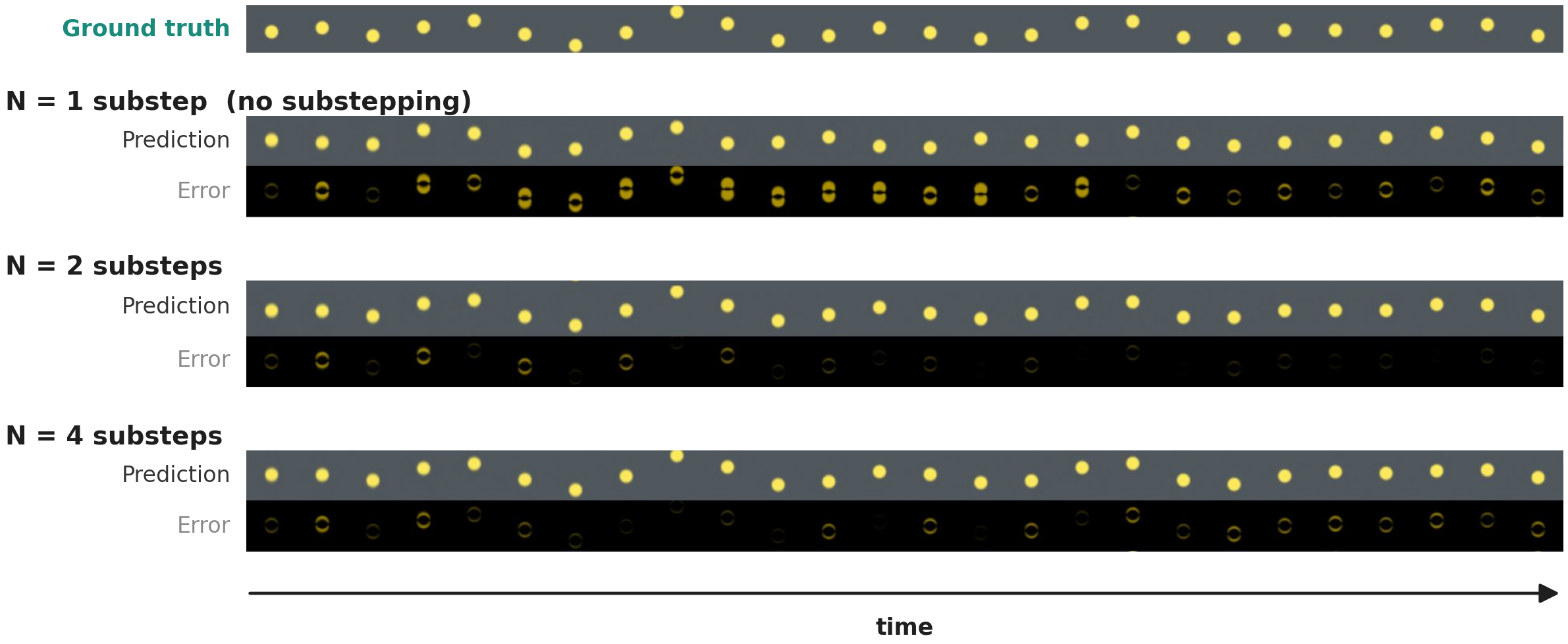}
\caption{Eval-time substepping recovers extrapolation on the oscillator. Top: ground-truth rollout. Each block below shows the spectrally-normed model's prediction (gray) and its error against the ground truth (black) at $N \in \{1, 2, 4\}$ integration substeps. At $N{=}1$ the prediction drifts out of phase and the error grows into bright double-lobed residuals. Refining the integration to $N{=}2$ and $N{=}4$ shrinks the error to faint traces, with the dynamics recovered once the effective step approaches the training scale.}
\label{fig:substepping}
\end{figure}

With the action force constrained, a qualitatively different failure remains. Above $\dt = 0.4$, (the training step size), the prediction no longer diverges in amplitude; instead it drifts out of phase with the ground truth, and the per-frame error grows into the bright double-lobed residuals of a position offset (Fig.~\ref{fig:substepping}, $N{=}1$). The model does not explode; it loses temporal alignment.
 
\begin{figure}[th]
\centering
\includegraphics[width=.8\textwidth]{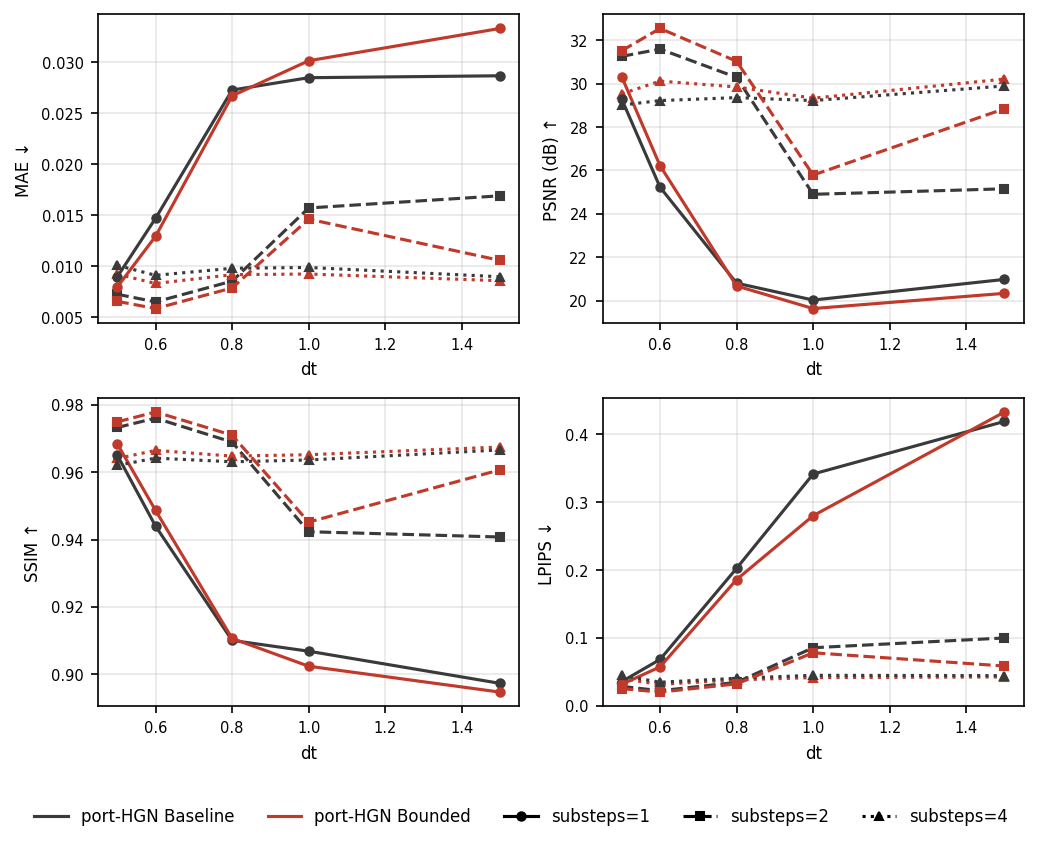}
\caption{Aggregate visual-quality metrics versus evaluation step size $\dteval$, restricted to step sizes beyond the training step ($\dttrain{=}0.4$), for the baseline port-HGN (black) and the spectrally-normed \emph{bounded} port-HGN (red). Line style encodes the number of inference-time integration substeps ($N{=}1$ solid, $N{=}2$ dashed, $N{=}4$ dotted). Without substepping ($N{=}1$) both models degrade steeply as $\dteval$ grows; increasing $N$ restores quality across MAE, PSNR, SSIM, and LPIPS, with the $N{=}4$ curves remaining high and nearly flat in $\dteval$. At intermediate substepping ($N{=}2$) the bounded model recovers more than the baseline.}
\label{fig:aggregate_metrics}
\end{figure}

The two candidate causes are a representational failure of the learned Hamiltonian beyond the training $\dt$ regime, or global truncation error accumulating in the fixed-step integrator. Inference-time substepping targets exclusively the second explanation, since it refines the integration without altering the learned dynamics or the inferred initial state. Refining the integration progressively removes the drift: at $N{=}2$ substeps the error collapses to faint traces and at $N{=}4$ it is almost entirely gone (Fig.~\ref{fig:substepping}), the same recovery that appears across all four aggregate metrics (Fig.~\ref{fig:aggregate_metrics}). The effective integration step at $\dt{=}1.5$ with $N{=}4$ is $1.5/4 \approx 0.375 \approx \dttrain$, and correct dynamics return when the integration step is brought back to the training scale. Because the learned dynamics are identical across these conditions, the recovery suggests that the failure is global truncation error rather than a deficiency of the learned Hamiltonian. In other words, when integrated at a step comparable to training, the same vector field reproduces correct dynamics at observation intervals well outside the training range.
 
\subsection{Recovering Both Directions}
\label{sec:recovery}

Taken together, the two fixes act on the two failure modes independently and combine into a single model that generalizes in both temporal directions. Figure~\ref{fig:aggregate_metrics} reports prediction quality across the extrapolation grid for the baseline and the spectrally-bounded port-HGN at $N \in \{1, 2, 4\}$ integration substeps. Without substepping ($N{=}1$) all four metrics degrade as $\dteval$ grows, most sharply the perceptual LPIPS distance, while SSIM and MAE change least in absolute terms, as expected for a scene with a large static background. Substepping recovers every metric monotonically, and by $N{=}4$ the curves are nearly flat in $\dteval$, the effective integration step having returned to the training scale. The bounded model recovers more completely than the baseline at intermediate substepping ($N{=}2$), indicating that constraining the forcing and refining the integration address complementary defects rather than the same one.

Because spectral normalization constrains only the action-force map and substepping is engaged only when the target step exceeds the training step, the two fixes address the extrapolation failures specifically. Interpolation below the training step is provided by the symplectic integrator, which refines any sub-training step without retraining (Fig.~\ref{fig:interp_extrap}), while subdividing large steps restores prediction above it (Fig.~\ref{fig:substepping}). A single port-Hamiltonian model trained at one frame rate therefore predicts at step sizes both finer and coarser than any it observed during training.

\section{Discussion}
\label{sec:discussion}

For deployment of Hamiltonian video models at variable frame rates, our results suggest a practical recipe: apply spectral normalization to any learned forcing term, and substep the integrator at inference time when the target playback rate exceeds the training rate. The computational cost of substepping is proportional to $N$ additional forward passes through the Hamiltonian network per observation step, and no retraining is required. The diagnosis attributes the two extrapolation failures to specific, individually removable causes: exponential magnitude growth to the unconstrained action-force map, and loss of temporal alignment to global truncation error in the fixed-step integrator.

\section{Conclusion}
\label{sec:conclusion}
 
A port-Hamiltonian video model grounds its predictions in a continuous-time energy function that is, in principle, queryable at any frame rate, and interpolation below the training step follows directly from its symplectic structure. In practice, however, this generalization breaks down once the rollout step grows beyond the training regime. We have shown that the breakdown is not a limitation of the learned representation but the product of two distinct numerical failure modes, namely latent magnitude growth driven by an unconstrained action-force map, and global truncation error accumulation from an under-resolved integrator. We provide a targeted fix for each, applying spectral normalization to the action-force map and substepping the integrator at inference time, and together they recover stable dynamics prediction at temporal resolutions well outside the training distribution, from a single model queried at rates both finer and coarser than the one it was trained on. These results recommend a practical recipe for enabling temporal generalization in continuous-time video generation.

\bibliographystyle{splncs04}
\bibliography{references}

\end{document}